\documentclass[conference]{IEEEtran}
\IEEEoverridecommandlockouts
\usepackage{cite}
\usepackage{amsmath,amssymb,amsfonts}
\usepackage{graphicx}
\usepackage{textcomp}
\usepackage{multirow}
\usepackage{algorithm}
\usepackage{algpseudocode}
\usepackage{url}
\usepackage{xcolor}
\def\BibTeX{{\rm B\kern-.05em{\sc i\kern-.025em b}\kern-.08em T\kern-.1667em\lower.7ex\hbox{E}\kern-.125emX}}
\begin{document}

\title{Assertion Detection in Clinical Natural Language Processing using Large Language Models}

\author{
\IEEEauthorblockN{1\textsuperscript{st} Yuelyu Ji}
\IEEEauthorblockA{\textit{Dept. of  Computing and Information} \\
\textit{University of Pittsburgh}\\
Pittsburgh, US  \\
yuj49@pitt.edu}
\and
\IEEEauthorblockN{2\textsuperscript{nd} Zeshui Yu}
\IEEEauthorblockA{\textit{Dept. of Pharmaceutical Sciences}\\
\textit{University of Pittsburgh}\\
Pittsburgh, US\\
zey1@pitt.edu}
\and
\IEEEauthorblockN{3\textsuperscript{rd} Yanshan Wang}
\IEEEauthorblockA{\textit{Dept. of Health Information Management} \\
\textit{University of Pittsburgh}\\
Pittsburgh, US \\
yanshan.wang@pitt.edu}
}

\maketitle

\begin{abstract}

In this study, we aim to address the task of assertion detection when extracting medical concepts from clinical notes, a key process in clinical natural language processing (NLP). Assertion detection in clinical NLP usually involves identifying assertion types for medical concepts in the clinical text, namely certainty (whether the medical concept is positive, negated, possible, or hypothetical), temporality (whether the medical concept is for present or the past history), and experiencer (whether the medical concept is described for the patient or a family member). These assertion types are essential for healthcare professionals to quickly and clearly understand the context of medical conditions from unstructured clinical texts, directly influencing the quality and outcomes of patient care. Although widely used, traditional methods, particularly rule-based NLP systems and machine learning or deep learning models, demand intensive manual efforts to create patterns and tend to overlook less common assertion types, leading to an incomplete understanding of the context. To address this challenge, our research introduces a novel methodology that utilizes Large Language Models (LLMs) pre-trained on a vast array of medical data for assertion detection. We enhanced the current method with advanced reasoning techniques, including Tree of Thought (ToT), Chain of Thought (CoT), and Self-Consistency (SC), and refine it further with Low-Rank Adaptation (LoRA) fine-tuning. We first evaluated the model on the i2b2 2010 assertion dataset. Our method achieved a micro-averaged F-1 of 0.89, with 0.11 improvements over the previous works. To further assess the generalizability of our approach, we extended our evaluation to a local dataset that focused on sleep concept extraction. Our approach achieved an F-1 of 0.74, which is 0.31 higher than the previous method. The results show that using LLMs is a viable option for assertion detection in clinical NLP and can potentially integrate with other LLM-based concept extraction models for clinical NLP tasks.

\end{abstract}

\begin{IEEEkeywords}
Assertion Detection Large Language Model In-context Learning LoRA Fine-tuning
\end{IEEEkeywords}

\section{Introduction}
\begin{figure}[htbp]
\centerline{\includegraphics[width=\linewidth]{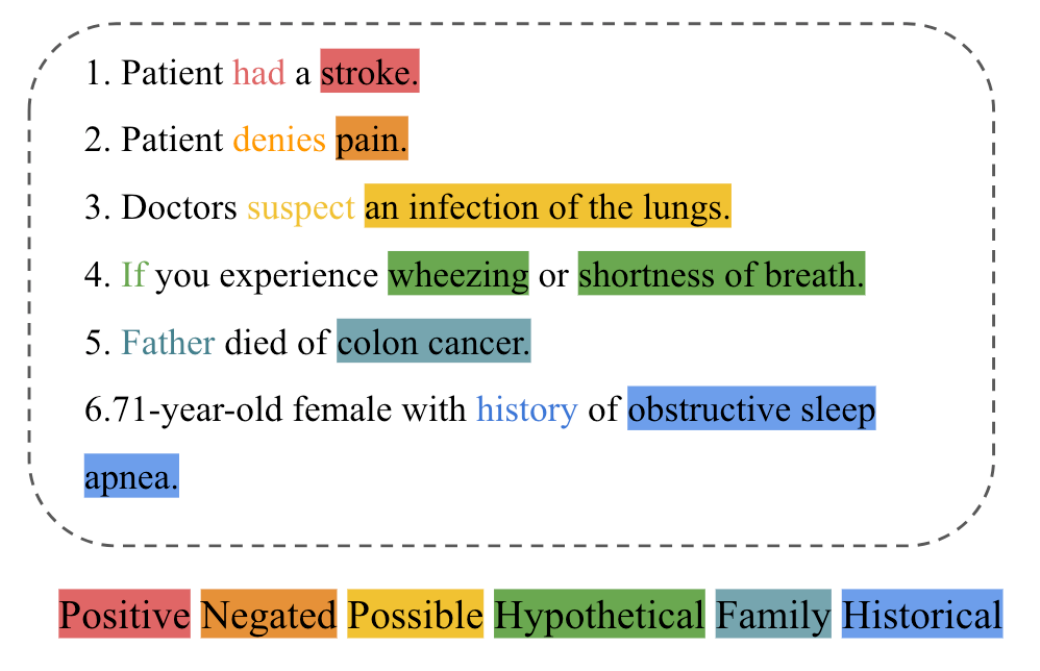}}
\caption{Examples of assertions in clinical texts. Medical concepts and the corresponding assertions are highlighted.}
\label{fig1}
\end{figure}

Assertion detection is a key task within the area of Clinical Natural Language Processing (NLP)\cite{harkema2009context}. It usually involves identifying the assertion types for medical concepts in the clinical text, namely certainty (whether the medical concept is positive, negated, possible, or hypothetical), temporality (whether the medical concept is for present or the previous history), and experiencer (whether the medical concept is described for the patient or a family member). Figure~\ref{fig1} shows an example of medical concepts and the corresponding assertions. This task plays a crucial role in understanding medical concepts from the free-text Electronic Health Records (EHRs), directly impacting the accuracy of clinical decision-making and the efficiency of patient care. As a core component of clinical NLP, assertion detection also holds significant potential for enhancing information retrieval and automated clinical reasoning. However, it faces challenges such as class distribution imbalance and the unstructured nature of clinical notes. Particularly challenging is the classification of assertions like 'Possible' and 'Family', which are often less frequently occurring and ambiguously expressed.
Previous studies have widely applied rule-based methods such as NegEx\cite{chapman2001simple} and ConText\cite{harkema2009context} in clinical NLP software, setting a benchmark in medical informatics with applications in tools like OHNLP Toolkit\cite{liu2023open}, MedTagger \cite{liu2013information}, medspaCy \cite{eyre2021launching}, and cTAKES \cite{savova2010mayo}. However, these rule-based approaches are limited by their fixed patterns and inability to exhaust all possibilities, often leading to low recall rates.
To overcome these limitations, deep learning methods like convolutional neural networks (CNNs) and Long short-term memory (LSTM) \cite{qian2016speculation,sergeeva2019negation,wang2022trustworthy} were introduced. Although these approaches show promise, they still require substantial amounts of labeled data and tend to underperform when dealing with small or imbalanced datasets.

To address the above limitations, recent attention has been focused on Large Language Models (LLMs) for their superior capacity to understand and generate human-like text. LLMs such as GPT-3\cite{brown2020language} and LLaMA\cite{touvron2023llama} are trained on vast datasets, enabling them to capture intricate patterns in language that rule-based systems cannot. Furthermore, these models introduce the concept of in-context learning \cite{olsson2022context}, an idea that enables LLMs to understand and perform new tasks efficiently by conditioning on a few examples in the input, thereby grasping the task's structure and generating response format through these illustrative examples. Our method employs in-context learning techniques, including Tree of Thought (ToT) \cite{yao2023tree}, Chain of Thought (CoT)\cite{wei2023chainofthought}, and Self-Consistency (SC)\cite{wang2023selfconsistency}. These methods leverage a small number of samples to rapidly understand new tasks and self-reflect, achieving meaningful success in clinical NLP tasks such as question answering and text generation\cite{arachchige2023large, yuan2023llm}.

In this paper, we treated assertion detection as a generative task, generating corresponding texts through various in-context learning methods, thereby utilizing the model's comprehension to tackle the task. Moreover, we introduced Low-Rank Adaptation (LoRA) fine-tuning \cite{hu2021lora} to enhance the LLM's understanding of instructions and achieved promising results with minimal data training.
We tested our in-context learning and LoRA fine-tuning techniques on two datasets, including the public i2b2 2010 assertion dataset\cite{uzuner20112010} and a local private corpus from the University of Pittsburgh Medical Center (UPMC). The validation on the local private corpus could validate the generalizability of the proposed approach. The results indicate that our method outperforms previous works across all six assertion categories on both datasets.

The three key contributions of this work are as follows:

\begin{itemize}
    \item We have developed and rigorously evaluated a range of LLMs enhanced by advanced reasoning methodologies, including ToT, CoT, and SC. These methods significantly improve the LLMs' capabilities in assertion detection, providing deeper insights and more trustworthy interpretations of medical narratives.

    \item Our study includes fine-tuning the LLaMA2-7B model\cite{touvron2023llama} using LoRA to achieve greater precision and contextual understanding. This optimization step refines the model's performance, making it more adept at handling the specific details of clinical assertion detection.\footnote{We open-source the code and models to the community for further research at: \url{https://github.com/JoyDajunSpaceCraft/Assertion_LLM}.} 

    \item The experiments on both public and private datasets show the fine-tuned LLMs' effectiveness in carefully detecting and categorizing medical assertions and highlight their generalizability and adaptability to specialized healthcare domains. 
\end{itemize}

\section{Related Work}

The landscape of clinical assertion detection has significantly evolved, with methodologies ranging from rule-based systems to advanced machine learning and deep learning approaches. 

One of the earliest and most influential methods is the ConText algorithm \cite{harkema2009context}, which utilizes hand-crafted patterns for negation and temporality classification in clinical text \cite{chapman2001simple}. ConText has been integral to various software applications, demonstrating its enduring impact on the field. Notably, it's been incorporated into the OHNLP Toolkit for EHR-based clinical research \cite{liu2023open}, MedTagger for cohort identification \cite{liu2013information}, medspaCy for clinical text processing \cite{eyre2021launching}, and cTAKES for text analysis and knowledge extraction \cite{savova2010mayo}. These implementations demonstrate ConText's significant influence and adaptability in medical informatics.

The i2b2 Challenge Assertions Task \cite{uzuner20112010} further motivated the development of machine learning models like SVMs and CRFs \cite{harkema2009context}, which provided improvements but also emphasized the complexity and challenges of clinical data.

The advent of deep learning introduced neural networks, including CNNs and LSTMs, to the task \cite{qian2016speculation,sergeeva2019negation}. These models showed promise but were often limited by their need for large labeled datasets. This limitation was solved to some extent by transformer-based models like NegBert \cite{DBLP:journals/corr/abs-1911-04211}, which marked significant advancements but still relied heavily on extensive labeled data.

Recently, prompt-based learning methods utilizing Large Language Models (LLMs) have emerged as a potent tool for clinical NLP tasks. These methods allow models to engage in few-shot learning and swiftly adapt to new tasks, as evidenced by \cite{wang2022trustworthy}. However, the literature on LLMs in clinical assertion detection has been scarce. Nonetheless, there is a paucity of literature regarding utilizing LLMs in clinical assertion detection. Our work expands upon this innovative foundation by amalgamating prompt engineering with sophisticated reasoning methodologies, such as ToT\cite{yao2023tree}, CoT\cite{wei2023chainofthought}, and SC \cite{wang2023selfconsistency}. These methods not only facilitate complex inference but also enhance the interpretability of language models, particularly in specialized tasks like medical diagnosis \cite{sivarajkumar2023empirical}.

Moreover, applying the LoRA method enables more efficient fine-tuning, thereby strengthening the model's capacity to generalize across a diverse range of clinical narratives \cite{gema2023parameter}. This approach addresses the previous limitations by reducing the dependency on large annotated datasets and improving model performance on minority classes.

The field has witnessed a wide array of methodologies, ranging from the rule-based approach of ConText to the most recent transformer-based models. Our research extends this evolution by harnessing advanced reasoning and efficient fine-tuning techniques in LLMs, thereby advancing the frontiers of clinical assertion detection.

\section{Dataset}

Our investigation into clinical assertion detection utilizes two datasets, namely the i2b2 assertion dataset and a local Sleep dataset, as listed in Table~\ref{tab:data_distribution_single_col}

\begin{table}[htbp]
\caption{Assertion Distribution in i2b2 and Sleep Datasets}
\begin{center}
\scalebox{0.8}{
\begin{tabular}{l|c|c|c|c}
\hline
\multirow{2}{*}{\textbf{Label}} & \multicolumn{2}{c|}{\textbf{i2b2}} & \multicolumn{2}{c}{\textbf{Sleep}} \\
\cline{2-5} 
 & \textbf{Train} & \textbf{Eval} & \textbf{Train} & \textbf{Eval} \\
\hline
Family & 185 (5.26\%) & 47 (7.79\%) & 40 (10.99\%) & 12 (13.04\%) \\
Historical &-& -& 81 (22.25\%) & 19 (20.65\%) \\
Hypothetical & 317 (9.02\%) & 48 (7.96\%) & 33 (9.07\%) & 9 (9.78\%) \\
Negated & 758 (21.57\%) & 127 (21.06\%) & 33 (9.07\%) & 15 (16.30\%) \\
Possible & 265 (7.29\%) & 42 (6.97\%) & 46 (12.64\%) & 9 (9.78\%) \\
Positive & 1988 (55.58\%) & 324 (53.73\%) & 131 (35.99\%) & 28 (30.43\%) \\
\hline
Total & 3513& 588& 364& 92\\
\hline
\end{tabular}
}
\label{tab:data_distribution_single_col}
\end{center}
\end{table}

\begin{figure*}[htbp]
\centerline{\includegraphics[width=180mm]{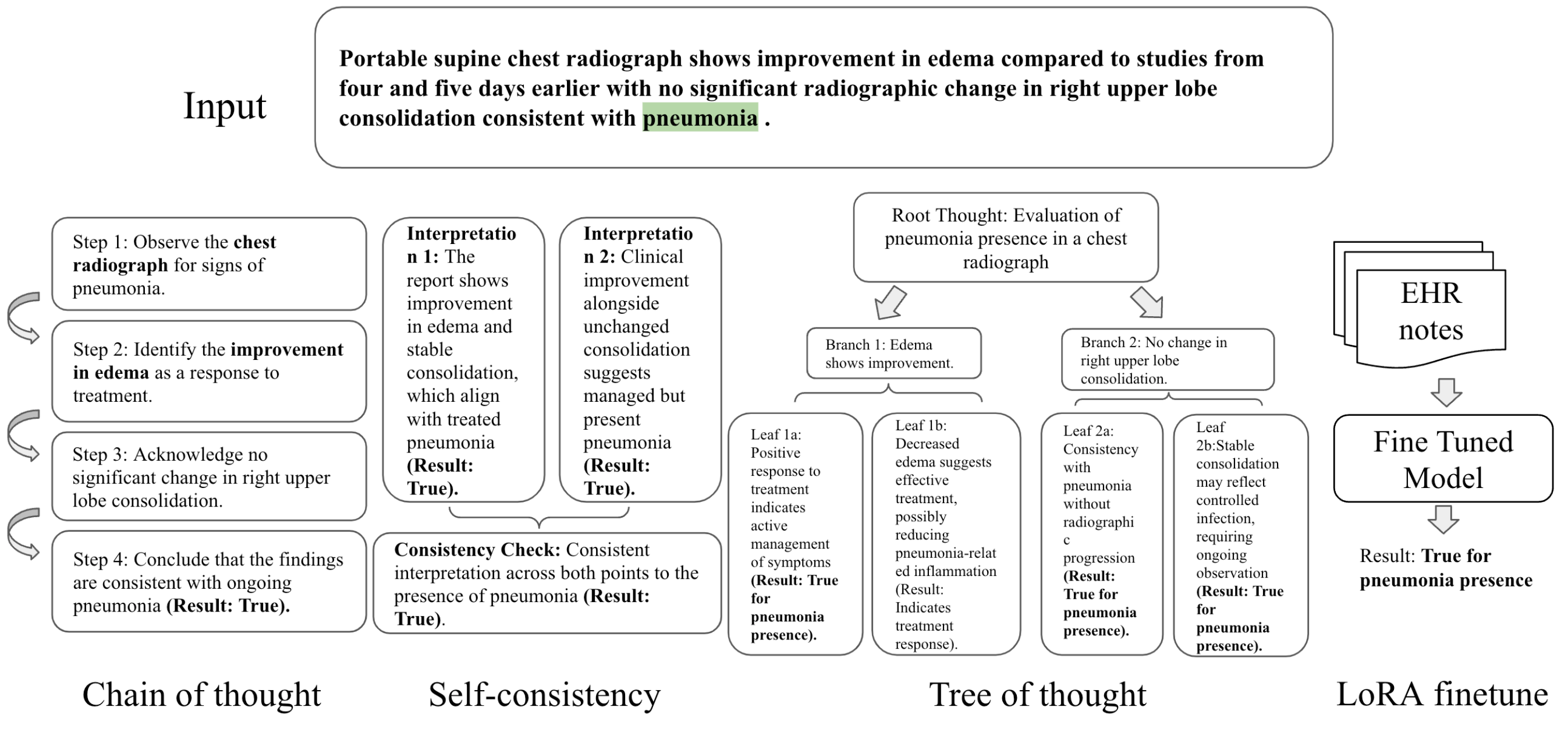}}
\caption{Methods used in the assertion detection.}
\label{fig2}
\end{figure*}

\textbf{1. i2b2 Dataset:} 
The i2b2 2010 assertion dataset provides annotated data from discharge summaries and progress notes sourced from three different medical institutions, as referenced in \cite{uzuner20112010}. It includes manual annotations for six types of assertions related to medical concepts within clinical documentation: Present (Positive), Absent (Negated), Possible, Hypothetical, Conditional, and Family. We only used the category matching our task to fit our assertion definition, so we didn't use the Conditional category in the i2b2 dataset.

\textbf{2. Sleep Dataset:} A curated dataset from UPMC comprises annotated records related to sleep disorders, specifically targeting snoring and obstructive sleep apnea (OSA).
This choice is motivated by the high prevalence of sleep disorders, their underdiagnosis, and the negative impact on quality of life, morbidity, and survival.

Sporadic snoring is often considered harmless, affecting a substantial proportion of the adult population (44 percent of males and 28 percent of females aged 30 to 60). However, persistent and loud snoring may suggest a potential underlying issue like OSA, warranting medical intervention. 
OSA is a sleep disorder characterized by obstructive apneas and hypopneas that occur when upper airway resistance is sufficient to disrupt sleep. Patients with OSA are at heightened risk of adverse clinical complications, such as metabolic syndrome, cardiovascular disease, or neuropsychiatric dysfunction \cite{kline2017clinical}.

The clinical notes were retrieved from the clinical data warehouse. We used keyword sampling to identify sleep-related notes and randomly sampled a total of 456 clinical notes. Annotations in this dataset reflect assertion types like Positive, Negated, Possible, Hypothetical, Family, and Historical. These granular categories enable an in-depth analysis of the LLMs' performance in recognizing and classifying specific medical assertions. The annotations were carefully reviewed and adjudicated by a healthcare professional (ZY, co-author of this study) to ensure their accuracy and reliability. This process involved a comprehensive examination of relevant clinical notes. This study is approved by The University of Pittsburgh’s Institutional Review Board (IRB).

The distribution for the data can be found in the Table \ref{tab:data_distribution_single_col}.

\section{Methodology}
Figure \ref{fig2} illustrates the proposed approach of using LLMs for assertion detection. 

\subsection{In-context Learning}
\subsubsection{Chain-of-Thought Prompting (CoT)}
Chain of Thought (CoT) prompting enhances the capability of language models to address complex tasks by breaking down problems into a series of granular and progressive subtasks. This approach leads to more structured and understandable solutions through methodical step-by-step reasoning. For example, when assessing a patient's symptoms, the reasoning might unfold as follows:

\begin{quote}
"The patient has been experiencing persistent migraines, often escalating to nausea in the evenings, which have not responded to over-the-counter medications for the last month."
\end{quote}
LLM will think stepwisely and generate the question based on understanding the case.

\textbf{Guided Question 1:} Considering the patient's description of their migraines as "persistent" and "escalating to nausea," can we assert that the condition is 'Positive' and currently affecting the patient's health?

\textbf{Expected Answer 1:} Yes, the description indicates an ongoing and troublesome condition, confirming a 'Positive' assertion for the current state of the patient's health.

\textbf{Guided Question 2:} Does the lack of response to over-the-counter medications over the last month strengthen the 'Positive' assertion for the severity and presence of the patient's condition?

\textbf{Expected Answer 2:} Yes, the fact that common treatments have been ineffective for a duration of "the last month" suggests that the condition is 'Positive' and may require further medical evaluation or treatment.
It will break down into different reasoning steps. By answering the questions step by step, LLM will provide the result of assertion detection.

\begin{table*}[htbp]
\caption{Performance Comparison Across Models and Methods on the i2b2 and Sleep Datasets.}
\begin{center}
\begin{tabular}{l|c|c|c|c|c|c|c|c|c|c|c|c}

\hline
\multirow{2}{*}{\textbf{Dataset}} & \multirow{2}{*}{\textbf{Assertion Category}} & \multicolumn{4}{c|}{\textbf{ChatGPT}} & \multicolumn{5}{c|}{\textbf{LLaMA2-7B}} & \multirow{2}{*}{BERT} & \multirow{2}{*}{ConText}\\
\cline{3-11} 
& & \textbf{Simple} & \textbf{CoT} & \textbf{ToT} & \textbf{SC} & \textbf{Simple} & \textbf{CoT} & \textbf{ToT} & \textbf{SC} & \textbf{LoRA} & \\
\hline
\multirow{6}{*}{\textbf{i2b2}}&Family & 0.67 & 0.7 & 0.55 & 0.57 & 0.87 & 0.85 & 0.85 & \textbf{0.92} & 0.67 & - & 0.72 \\
&Historical & - & - & - & - & - & - & - & - & - & - & - \\
&Hypothetical & 0.66 & 0.56 & 0.68 & 0.55 & 0.94 & 0.91 & 0.91 & \textbf{0.96} & 0.875 & - & - \\
&Negated & 0.53 & 0.57 & 0.55 & 0.69 & 0.86 & 0.88 & 0.9 & 0.93 &\textbf{0.98} & 0.84 & 0.74 \\
&Possible & 0.63 & 0.66 & 0.65 & 0.7 & 0.95 & 0.95 & 0.93 & 0.95 &\textbf{ 0.96} & 0.0 & 0.0 \\
&Positive & 0.62 & 0.66 & 0.65 & 0.72 & 0.88 & 0.9 & 0.91 & 0.95 & \textbf{0.99} & 0.81 & 0.89 \\

\hline
\multirow{6}{*}{\textbf{Sleep}}&Family & \textbf{0.72} & 0.5 & 0.22 & 0.43 & 0.53 & 0.55 & 0.46 & 0.4 & 0.53 & - & 0.6 \\
&Historical & 0.72 & 0.63 & 0.61 & 0.67 & 0.76 & 0.86 & \textbf{0.9} & 0.71 & 0.76 & - & 0.7 \\
&Hypothetical & 0.44 & 0.55 & 0.11 & 0.44 & 0.11 & 0.11 & 0.11 & 0.0 & \textbf{0.88} & - & - \\
&Negated & 0.4 & 0.36 & \textbf{0.5} & 0.0 & 0.14 & 0.5 & 0.27 & 0.29 & 0.14 & 0.25 & 0.3 \\
&Possible & 0.0 & 0.29 &\textbf{0.57} & 0.33 & 0.36 & 0.36 & 0.36 & 0.5 & 0.36 & 0.0 & 0.0 \\
&Positive & 0.62 & 0.78 & 0.74 & 0.69 & 0.91 & 0.85 & 0.83 & 0.83 & \textbf{0.92} & 0.46 & 0.58 \\
\hline
\end{tabular}
\label{tab:sleep_dataset_comprehensive_performance}
\end{center}
\end{table*}

\subsubsection{Self-Consistency over Diverse Reasoning Paths (SC)}
SC over Diverse Reasoning Paths evaluates the consistency of multiple reasoning pathways, favoring the most common conclusion. It optimizes the robustness of decision-making processing by considering various possible solutions and selecting the most frequent one, thereby reducing reliance on a single CoT. 
Formalizing the self-consistency is captured by the equation:
\begin{equation}
   \hat{a} = \arg\max_{a \in A} \sum_{i=1}^{m} (a_i = a)
\end{equation}

The $\hat{a}$ is the final assertion outcome. $A$, which in the context of medical assertion detection typically includes 'True' or 'False'.  $m$ is the number of interpretative paths generated, which in Figure \ref{fig2} self-consistency is 2. When the assertion outcome 
$a_i$ of the $i_{th}$ path aligns with the assertion $a$. The method aggregates the outcomes of all paths and selects the most frequently occurring assertion as the final diagnosis.
\subsubsection{Tree of Thoughts (ToT) Framework}
The ToT framework employs a heuristic-guided decision tree to optimize problem-solving. It decomposes complex problems, explores multiple reasoning paths, and uses heuristics to determine the most effective solution. And for every part, LLM would solve the partial problems.

Formalizing the optimal reasoning path is captured by the equation:
\begin{equation}
s^* = \arg\max_{s \in S} V(s),
\end{equation}
where \( s^* \) is the most supported conclusion, \( S \) is the set of reasoning paths considered, and \( V(s) \) represents the heuristic evaluation of each path. In the context of the medical diagnosis depicted, \( s^* \) is the diagnosis affirmed by the process, \( S \) includes the LLM's observations and their interpretations, and \( V(s) \) reflects the evaluation  of these hypotheses based on their alignment with known clinical information.
At the same time, it is essential to ensure that the chosen path in the search process is optimal within the ToT framework. This guarantees that our model navigates through the reasoning space efficiently.
\subsection{Efficient Fine-Tuning Method}
For the specific task of clinical assertion detection, we efficiently fine-tuned our pre-trained language model, LLaMA2-7B with LoRA. LoRA introduces trainable low-rank matrices \(A\) and \(B\) into the Transformer layers by reducing the number of parameters requiring adaptation. This technique allows for quicker adaptation and resource-efficient customization, which is particularly beneficial for our medical-focused model that demands high precision and interpretability. 

The adaptation process is formulated in the following equation:
\begin{equation}
    \Delta W = BA
\end{equation}
where \(\Delta W\) is the update to the pre-trained weight matrix \(W\). This reparameterization strategy enables targeted fine-tuning, which is crucial for processing complex clinical narratives for the i2b2 dataset and extracting sleep-related conditions for the Sleep dataset.

\section{Experiments and Results}

\subsection{Experimental Setup}

We evaluated the performance of two LLMs, LLaMA2-7B, and ChatGPT 3.5 turbo, across different assertion categories using the F-1 score metric. We employed CoT, ToT, and SC prompt engineering approaches for both LLMs. In addition, the open source LLM LLaMA2-7B was fine-tuned with LoRA on an NVIDIA A100 GPU,  with each session lasting one hour. Due to privacy considerations, we used the Azure OpenAI ChatGPT 3.5 turbo. In comparison, the BERT model \cite{van-aken-etal-2021-assertion} and ConText \cite{harkema2009context} algorithm are used as baseline approaches. The Simple approach used no in-context learning method and only asked the LLM whether the medical term categorized to the assertion.

\subsection{Dataset and Methodological Comparative Analysis}
 Our analysis across the i2b2 and Sleep datasets(refer to Tables \ref{tab:sleep_dataset_comprehensive_performance}) showed notable F1 score variability across different datasets and assertion categories. 
 
\textbf{Performance Comparison Across Assertion Categories:}
We observed that optimal F1 scores vary across different assertion categories on the two datasets. Specifically, the 'Family,' 'Hypothetical,' and 'Positive' categories demonstrated relatively minor disparities in their best F1 scores across the datasets: 0.72 and 0.92 for the 'Family,' 0.96 and 0.88 for the 'Hypothetical,' and 0.99 and 0.92 for the 'Positive' category, respectively. The close of these scores suggested that the recognition performance for these two categories was comparatively stable across datasets. Conversely, the 'Historical', 'Hypothetical', 'Negated', and 'Possible' categories exhibited significant differences in their best F1 scores between the datasets. For example, the best F1 score across datasets was 0.98 and 0.5 for 'Negated' and 0.95 and 0.57 for 'Possible'.

\textbf{Performance Comparison Across Approaches:}
Our examination of the i2b2 and Sleep datasets revealed distinct patterns in model performance. In the i2b2 dataset, methods like Simple, CoT, ToT, and SC consistently delivered strong results across categories, with LLaMA2-7B performing better in in-context learning. Conversely, in the Sleep dataset, ChatGPT's ToT method excelled in 'Negated' (0.5) and 'Possible' (0.57) categories.

Notably, the LLaMA2-7B model, when paired with the ToT approach, achieved an exceptional F1 score of 0.90 in 'Historical' within the Sleep dataset. Moreover, the fine-tuned LoRA technique performed better in the i2b2 dataset, particularly in 'Negated' (0.98) and 'Positive' (0.99), and continued to perform well in the Sleep dataset, especially in 'Hypothetical' (0.88) and 'Positive' (0.92).

Comparatively, baseline models like BERT and ConText also showed commendable results. For instance, ConText's 'Family' score (0.72) outperformed ChatGPT's CoT in the i2b2 dataset, and BERT's performance in 'Negated' (0.84) outdid ChatGPT's CoT, emphasizing the competitive edge of traditional models in certain contexts.

\textbf{Performance Comparison Across Dataset:}
Analysis of the i2b2 and Sleep datasets revealed notable variances in F1 scores across assertion categories.'Hypothetical' and 'Positive' categories showed relatively stable performance, with 'Hypothetical' scoring 0.88 and 0.96 and 'Positive' scoring 0.92 and 0.99 across Sleep and i2b2 datasets respectively 'Family' presented minor variation, scoring 0.72 in Sleep and 0.92 in i2b2. Conversely, the 'Negated' and 'Possible' categories showed more significant variations. For instance, in the 'Negated' category, the range was from 0.5 in the Sleep dataset to 0.98 in the i2b2 dataset. Similarly, in the 'Possible' category, it varied from 0.57 in Sleep to 0.96 in i2b2. These disparities emphasize the complexity of achieving consistent recognition across different datasets.

\section{Discussion}
\subsection{Error Analysis}
We conducted an error analysis of the proposed approaches on two datasets.
\subsubsection{Contextual Ambiguity}
Contextual ambiguity presents a significant challenge in clinical assertion detection. It occurs when models encounter clinical narratives with symptoms or conditions described ambiguously or indirectly. For example, terms like 'snoring' and 'sleepiness during the day' in clinical narratives (Table \ref{tab:clinical_case_study}) might suggest various conditions beyond OSA. However, in the absence of a clear diagnostic label, models may mistakenly think this description indicates OSA. This results in wrong, where the model incorrectly labels a case as 'Positive' for OSA without a definitive present diagnosis, as demonstrated in the clinical narrative below:

\begin{table}[htbp]
\caption{Clinical Narrative Demonstrating Contextual Ambiguity.}
\begin{center}
\begin{tabular}{|l|}
\hline
Easy bruising sleep: \underline{snoring}, \underline{sleepiness during the day} , \\
 \underline{falling asleep at work}, \underline{falling asleep easily} 
watching tv, insomnia \\psych: depression, anxiety, 
panic attacks,
suicidal thoughts, homicidal\\ thoughts alcoholic drinks per day: 0 days per week: I have personally \\reviewed the  nursing note and triage note for this patient. \\
\hline
\textbf{Model Output:} 'Positive' for OSA \\
\hline
\textbf{Correct Label:} Not Specified \\
\hline
\end{tabular}
\label{tab:clinical_case_study}
\end{center}
\end{table}

\subsubsection{Long Dependencies}

Long dependencies are characterized by the need to connect entities and their associated cues that are separated by more than a few tokens. The example is shown in Table \ref{tab:long_dependency_example}. These dependencies are responsible for a substantial portion of errors, as models might not be configured to consider distant assertion cues effectively. 
\begin{table}[htbp]
\caption{Clinical Narrative Demonstrating Long Dependency}
\begin{center}
\begin{tabular}{|l|}
\hline
After an initial assessment of \underline{joint stiffness}, the patient's\\ symptoms were managed conservatively.\ldots Further evaluation\\ revealed  \underline{significant improvement}, and the patient's \\
\underline{previously reported symptoms  are no longer present}. \\
\hline
\textbf{Model Output:} 'Positive' for joint stiffness \\
\hline
\textbf{Correct Label:} 'Negated' for joint stiffness \\
\hline
\end{tabular}
\label{tab:long_dependency_example}
\end{center}
\end{table}
Here, the resolution of symptoms away from a distant cue should negate the initial assertion of the condition, but if the model focuses only on adjacent information, this key information might be missed. 

\subsection{Limitation}
This study is subject to several limitations. Firstly, the generalizability of the study is confined, as the clinical note data was sourced solely from one institution. Expanding the dataset to include clinical notes from multiple institutions might enhance the robustness of the study. Secondly, we employed the LLaMA2-7B model due to computational constraints. While acknowledging that larger models, such as the 13B and 70B variants, might offer improved results, their usage was impractical within our computing framework. Future studies utilizing these larger models could uncover additional performance enhancements. Thirdly, the domain of LLMs is evolving rapidly. 

\section{Future work}
Our results suggest that while traditional tools like ConText have promising performance, LLMs and new  methods such as ToT, CoT, and SC offer significant improvements in analyzing complex clinical texts. Given the prevalent application of ConText in diverse clinical NLP tools, LLMs have the potential to supplant older rule-based approaches effectively. This transition heralds a promising avenue for future investigations in clinical assertion detection and healthcare analytics, potentially culminating in more dependable and efficient results.

\section*{Acknowledgments}

This work was supported by the National Institutes of Health under award number U24 TR004111 and R01 LM014306. The content is solely the responsibility of the authors and does not necessarily represent the official views of the National Institutes of Health. We extend our gratitude to Dr. Zeshui Yu for his diligent work in annotating the dataset.

\bibliographystyle{IEEEtran}
\bibliography{print_ready}
\end{document}